\documentclass[10pt,twocolumn,letterpaper]{article}

\usepackage{wacv}
\usepackage{times}
\usepackage{epsfig}
\usepackage{graphicx}
\usepackage{amsmath}
\usepackage{amssymb}

\usepackage{subcaption}
\usepackage{array}

\usepackage{mathtools}
\usepackage{amsmath}
\usepackage{bbm}
\usepackage{multirow}
\usepackage{wrapfig}

\usepackage{color} % jhchoi @ 2012.6.29
\usepackage[pagebackref=true,breaklinks=true,letterpaper=true,colorlinks,bookmarks=false]{hyperref}

\wacvfinalcopy % *** Uncomment this line for the final submission

\def\wacvPaperID{372} % *** Enter the wacv Paper ID here

\ifwacvfinal\fi
\begin{document}

%%%%%%%%% TITLE
\title{ActionFlowNet: Learning Motion Representation for Action Recognition}
\author{
  Joe Yue-Hei Ng$^1$ \hspace{1em}
  Jonghyun Choi$^2$ \hspace{1em}
  Jan Neumann$^3$ \hspace{1em}
  Larry S. Davis$^1$ \vspace{.3em}
  \\
  $^1$University of Maryland, College Park \hspace{1.5em}
  $^2$Allen Institute for AI \hspace{1.5em}
  $^3$Comcast Labs, DC \vspace{.3em}\\
{\tt\small \{yhng,lsd\}@umiacs.umd.edu \hspace{1em}
jonghyunc@allenai.org \hspace{1em}
jan\_neumann@cable.comcast.com }
}

\maketitle
\ifwacvfinal\thispagestyle{empty}\fi

%%%%%%%%% ABSTRACT
\begin{abstract}
%!TEX root = ./wacv2018_flowaction.tex

%Even with the recent advances in convolutional neural networks (CNN) in various visual recognition tasks, the state-of-the-art action recognition system still relies on hand crafted motion feature such as optical flow to achieve the best performance.
We present a data-efficient representation learning approach to learn video representation with small amount of labeled data.
We propose a multitask learning model ActionFlowNet to train a single stream network directly from raw pixels to jointly estimate optical flow while recognizing actions with convolutional neural networks, capturing both appearance and motion in a single model.
%We additionally provide insights to how the quality of the learned optical flow affects the action recognition.
Our model effectively learns video representation from motion information on unlabeled videos.
Our model significantly improves action recognition accuracy by a large margin $(23.6\%)$ compared to state-of-the-art CNN-based unsupervised representation learning methods trained without external large scale data and additional optical flow input.
Without pretraining on large external labeled datasets, our model, by well exploiting the motion information, achieves competitive recognition accuracy to the models trained with large labeled datasets such as ImageNet and Sport-1M.

%Classification loss on video frames may not be the best supervision for learning motions but the appearances of the objects and their contexts. 
%since it is evident that training a classifier on both appearance of objects and a short motion information such as optical flow significantly helps action recognition even without a large dataset.
%To incorporate both appearance of objects and a short motion information, two stream approaches that take both raw frames and optical flow are widely used to improve the action recognition accuracy.
%Computing the high quality optical flow, however, is still an open problem and is not efficient in test time as the model requires to run an optical flow algorithm.
%We propose to train a single stream feed-forward multitask convolutional neural network with a small dataset for action recognition jointly with estimating the optical flow.
%Our model achieves comparable recognition accuracy to the state-of-the-art frame-based action classification model only without pre-training with large datasets and no optical flow computed in advance, and shows comparable end-point error in optical flow estimation to the state-of-the-art optical flow algorithm.

\end{abstract}

%%%%%%%%% BODY TEXT
%!TEX root = ./wacv2018_flowaction.tex

\section{Introduction}
\label{sec:intro}

Convolutional Neural Networks have demonstrated great success to multiple visual recognition tasks.
With the help of large amount of annotated data like ImageNet, the network learns multiple layers of complex visual features directly from raw pixels in an end-to-end manner without relying on hand-crafted features. 
Unlike image labeling, manual video annotation often involves frame-by-frame inspection and temporal trimming of videos that are expensive and time consuming. 
This prohibits the technique to be applied to other problem domains like medical imaging where data collection is difficult.

We focus on effectively learning video motion representation for action recognition without large amount of external annotated video data.
Following previous work~\cite{misra2016shuffle,vondrick2016generating,fernando2016self} that leverages spatio-temporal structure in videos for unsupervised or self-supervised representation learning, we are interested in learning video representation from motion information encoded in videos in addition to semantic labels.

\begin{figure}[h!]
  \vspace{-1em}
\centering
	\includegraphics[width=1.00\linewidth]{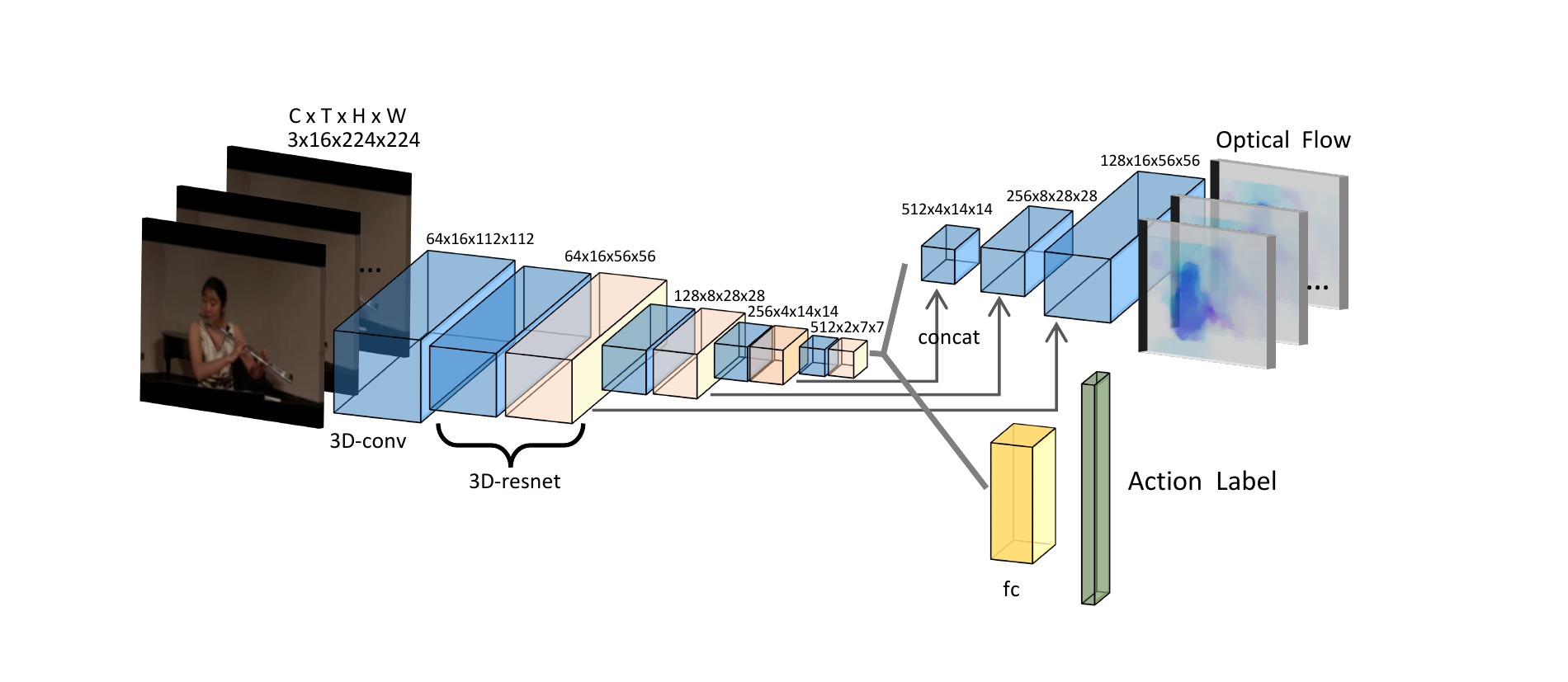}	
  \caption{ActionFlowNet for jointly estaimting optical flow and recognizing actions. Orange and blue blocks represent ResNet modules, where blue blocks represents strided convolution. Channel dimension is not shown in the figure.}
	\label{fig:vid_afn}
  \vspace{-1em}
\end{figure}

Learning motion representation on videos from raw pixels is challenging.
With large scale datasets such as Sports-1M~\cite{karpathyTSLSF14} and Kinetics~\cite{kay2017kinetics}, one could train a high capacity classifier to learn complex motion signatures for action recognition by extending image based CNN architectures with 3D convolutions for video action recognition~\cite{karpathyTSLSF14,tranBFTP15,carreira2017quo}.
However, while classification loss is an excellent generic appearance learner for image classification, it is not necessarily the most effective supervision for learning motion features for action recognition.
As shown in~\cite{carreira2017quo}, even with large amount of labeled video data, the model still benefits from additional optical flow input stream. This suggests that the model is ineffective in learning motion representation for action recognition from video frames, and thus alternative approach should be explored for learning video representation.

Two-stream convolutional neural networks, which separately learn appearance and motion by two convolutional networks on static images and optical flow respectively, show impressive results on action recognition~\cite{simonyanZ14a}. 
The separation, however, fails to learn the interaction between the motion and the appearance of objects, and introduces additional complexity of computing the flow to the classification pipeline.
In addition, human visual system does not take optical flow as front end input signals but infer the motion from raw intensities internally.
Therefore, we focus to learn both motion features and appearance directly from raw pixels without hand-crafted flow input.

Encouraged by the success on estimating optical flow with convolutional neural networks~\cite{FischerDIHHGSCB15}, we train a single stream feed-forward convolutional neural network - \mbox{\emph{ActionFlowNet}} - for jointly recognizing actions and estimating optical flow.
Specifically, we formulate the learning problem as multitask learning, which enables the network to learn both appearance and motion in a single network from raw pixels.
The proposed architecture is illustrated in Figure~\ref{fig:vid_afn}.
With the auxiliary task of optical flow learning, the network effectively learns useful representations from motion modeling without a large amount of human annotation.
Based on the already learned motion modeling, the model then only requires action annotations as supervision to learn action class specific details, which results in requiring less annotation to perform well for action recognition.

Our experiments and analyses show that our model successfully learns motion features for action recognition and provide insights on how the learned optical flow quality affects action classification.
We demonstrate the effectiveness of our learned motion representation on two standard action recognition benchmarks - UCF101 and HMDB51.
Without providing external training data or fine-tuning from already well-trained models with millions of samples, we show that jointly learning action and optical flow significantly boosts action recognition accuracy compared to state-of-the-art representation learning methods trained without external labeled data.
Remarkably, our model outperforms the models trained with large datasets Sports-1M pretrained C3D by 1.6\% on UCF101 dataset, showing the importance of feature learning algorithms.

%!TEX root = ./cvpr2017_flowaction.tex

\section{Related Work}
\label{sec:related}
Over the past few years, action recognition accuracy has been greatly improved by learned features and various learning models utilizing deep networks.
%Please refer to a comprehensive survey of recent advances in action recognition by Cheng \etal~\cite{chengWSNB15}.
% deep model for action
%We utilize the deep model for action recognition.
Two-stream network architecture was proposed to recognize action using both appearance and motions separately~\cite{simonyanZ14a}.
A number of follow up methods have been proposed based on two-stream networks that further improved action recognition accuracies~\cite{feichtenhofer2016convolutional,wang2016actions,wangXWQLTV16,feichtenhofer2016spatiotemporal,ng2018tdn}.
%TODO: add my own citation there
Our work is motivated by their success in incorporating optical flow for action recognition, but we focus on learning from raw pixels instead of relying on hand-crafted representations.

%Ng \etal developed network architectures using feature pooling and the long short term memory (LSTM) module and trained with very long video clips~\cite{ngHVVMT15}.
%In contrast, we focus on learning motion features from short duration clips.
%The long term information aggregation techniques are complementary to ours and could further improve accuracy when integrated.

% optical flow
Optical flow encodes motion between frames and is highly related to action recognition.
%EpicFlow~\cite{ravaudWHS15} is one of the best performing optical flow estimation algorithms.
Our model is motivated by the success of FlowNet~\cite{FischerDIHHGSCB15} and 3D convolutions for optical flow estimation in videos~\cite{tranbftp16}, but emphasizes on improving action recognition.

% action with pre-trained models
Pre-training the network with a large dataset helps to learn appearance signatures for action recognition.
Karpathy \etal proposed a ``Slow Fusion'' network for large scale video classification~\cite{karpathyTSLSF14}.
Tran \etal trained a 3D convolutional neural network (C3D) with a large amount of data and showed the learned features are generic for different tasks~\cite{tranBFTP15}.
Recently, Carreira and Zisserman trained I3D models~\cite{carreira2017quo} on the Kinetics dataset~\cite{kay2017kinetics} and achieved strong action recognition performance.
In contrast, since training networks on such large scale datasets is extremely computationally expensive, we focus on learning from small amounts of labeled data.
With only small amount of labeled data, we show that our model performs competitive to models trained with large datasets.

Leveraging videos as a source for unsupervised learning has been suggested to learn video representations without large labeled data.
Different surrogate tasks have been proposed to learn visual representations from videos without any labels.
Wang \etal trained a network to learn visual similarity for patches obtained from visual tracking in videos~\cite{wang2015unsupervised}.
Misra \etal trained a network to differentiate the temporal order of different frames from a video~\cite{misra2016shuffle}.
Jacob \etal learned apperance features by predicting the future trajectories in videos~\cite{walker2016uncertain}.
Fernando \etal proposed Odd-One-Out networks (O3N) to identify video sequences that are out of order for self-supervised learning~\cite{fernando2016self}.
Our work, similarly, uses video as an additional source for learning visual representation.
However, in contrast to previous work which focused on learning visual representations for a single image, we learn motion representations for videos which models more than a single frame.
Vondrick \etal used a Generatie Adversarial Network to learn a generative model for video~\cite{vondrick2016generating}.
We focus on learning motion representations but not video generation.

Independent to our work, Diba \etal trained a two stream network with flow estimation~\cite{diba2016efficient}.
They based their network on C3D with a two-stream architecture.
Our work employs a single stream network to learn both appearance and motion.
While we both estimate motion and recognize actions in the same model, we focus on learning motion representations without pretraining on large labeled datasets and provide more analysis to learn flow representations for action recognition.

%!TEX root = ./wacv2018_flowaction.tex

\section{Approach}
\label{sec:approach}

We propose a single end-to-end model to learn both motions and action classes simultaneously.
Our primary goal is to improve action classification accuracy with the help of motion information; we use optical flow as a motion signature.
Unlike previous methods that utilize externally computed optical flow as the input to their models, we only use the video frames for input and simultaneously learn the flow and class labels.

\subsection{Multi-frame Optical Flow with 3D-ResNet}
%Learning from more frames is known to improve recognition performance~\cite{ngHVVMT15,wang2016temporal}. We train multi-frame models which learn motion from multiple frames for better action recognition. 

%%%%%%%%%%%%%%%%%%%%%%%%%%%%%%%%%%%%%%%%%%%%
%\textbf{Learning Flows with Residual Network.}
Fischer \etal proposed FlowNet~\cite{FischerDIHHGSCB15} that is based on convolutional neural networks to estimate high quality optical flow. Tran \etal proposed to use 3D convolution and deconvolution layers to learn multi-frame optical flow from videos~\cite{tranbftp16}.
In addition, He \etal introduced residual networks (ResNet) to train a deeper convolutional neural network model by adding shortcut connections~\cite{heZRS15}.

In addition to the benefit of easy training, ResNet is fully convolutional, so is easily applied to pixel-wise prediction of optical flow, unlike many architectures with fully connected layers including AlexNet~\cite{krizhevskySH12} and VGG-16~\cite{simonyanZ14}.
In contrast to other classification architectures like AlexNet and VGG-16, which contains multiple max pooling layers that may harm optical flow estimation, the ResNet architecture only contains one pooling layer right after conv1.
We believe the reduced number of pooling layers makes ResNet more suitable for optical flow estimation where spatial details need to be preserved.
Specifically, we use an 18 layers ResNet, which is computationally efficient with good classification performance~\cite{heZRS15}.

Taking advantage of ResNet for flow estimation, we extend ResNet-18 to \emph{3D-ResNet-18} for multi-frame optical flow estimation by replacing all $k\times k$ 2D convolutional kernels with extra temporal dimension $k \times k \times 3$, inspired by~\cite{tranbftp16}.
The deconvolution layers in the decoder are extended similarly.
Skip connections from encoder to decoder are retained as in~\cite{FischerDIHHGSCB15} to obtain higher resolution information in the decoder.
Unlike \cite{FischerDIHHGSCB15}, we only use the loss on the highest resolution to avoid downsampling in the temporal dimension. We do not apply temporal max pooling suggested in~\cite{tranBFTP15,tranbftp16}, but use only strided convolutions to preserve temporal details. After the third residual block, the temporal resolution is reduced by half when the spatial resolution is reduced.

\textbf{Future Prediction.}
In addition to computing the optical flow between the $T$ input frames, we train the model to predict the optical flow on the last frame, which is the optical flow between the $T^{th}$ and $(T+1)^{st}$ frames.
There are two benefits of training the model to predict the optical flow of the last frame:
1) It is practically easier to implement a model with the same input and output sizes, since the output sizes of deconvolution layers are usually multiples of the inputs; and
2) Semantic reasoning is required for the model to extrapolate the future optical flow given the previous frames. This possibly trains the model to learn better motion features for action recognition, as also suggested by previous work ~\cite{walker2016uncertain}, which learned appearance feature by predicting the future.

Following~\cite{FischerDIHHGSCB15}, the network is optimized over the end-point error (EPE), which is the sum of $L_2$ distance between the ground truth optical flow and the obtained flow over all pixels.
The total loss for the multiple frame optical flow model is the EPE of $T$ output optical flow frames:
\vspace{-.5em}
\begin{equation}
  \sum_{t=1}^T\sum_{p} \|\mathbf{o}_{j,t,p} - \widehat{\mathbf{o}_{j,t,p}} \|_2,
	\label{eq:loss_epe}
  \vspace{-.5em}
\end{equation}
where $\mathbf{o}_{j,t,p}$ is 2-dimensional optical flow vector of the $t^{th}$ and the $(t+1)^{st}$ frame in the $j^{th}$ video at pixel $p$.

Note that the $T^{th}$ optical flow frame $\mathbf{o}_{j,t}$ is the future optical flow for the $T^{th}$ and $(T+1)^{st}$ input frames, where the $(T+1)^{st}$ frame is not given to the model.

\subsection{ActionFlowNet}
\label{sec:multi}
\noindent \textbf{Knowledge Transfer by Finetuning.}
Finetuning a pretrained network is a common practice to transfer knowledge from different datasets and tasks.
Unlike previous work, where knowledge transfer has been accomplished between very similar tasks (image classification and detection or semantic segmentation), knowledge transfer in our model is challenging since the goals of pixel-wise optical flow and action classification are not obviously compatible. 
We transfer the learned motion by initializing the classification network using a network trained for optical flow estimation.
Since the network was trained to predict optical flow, it should encode motion information in intermediate levels which support action classification.
However, finetuning a pretrained network is known to have the problem of catastrophic forgetting.
Specifically, when training the network for action recognition, the originally initialized flow information could be destroyed when the network adapts the appearance information.
We prevent catastrophic forgetting by using the multitask learning framework.

\noindent \textbf{ActionFlowNet.}
To force the model to learn motion features while training for action recognition, we propose a multitask model \emph{ActionFlowNet}, which simultaneously learns to estimate optical flow, together with predicting the future optical flow of the last frame, and action classification to avoid catastrophic forgetting.
With optical flow as supervision, the model can effectively learn motion features while not relying on explicit optical flow computation.

In our implementation, we take 16 consecutive frames as input to our model.
In the last layer of the encoder, global average pooling across the spatial-temporal feature map, with size $512 \times 2\times 7 \times 7$, is employed to obtain a single 512 dimensional feature vector, followed by a linear softmax classifier for action recognition.  The architecture is illustrated in Figure~\ref{fig:vid_afn}.
The multitask loss is given as follows:
\vspace{-1em}
\begin{multline} 
%\resizebox{8cm}{!}{$
    \text{MT-Loss}_{j} = \underbrace{-\mathbbm{1}(y_j = \widehat{y_j})\log p(\widehat{y_j})}_{\text{Classification Loss}}~~ +\\
    ~~\lambda \underbrace{\sum_{t=1}^T\sum_{p} \|\mathbf{o}_{j,t,p} - \widehat{\mathbf{o}_{j,t,p}} \|_2}_{\text{Flow Loss}},
%$}
\label{eq:multi_mt_loss}
  \vspace{-1em}
\end{multline}
where $\mathbbm{1}(\cdot)$ is a indicator function, $y_j$ and $\widehat{y_j}$ are the groundtruth and predicted action labels respectively of the $j^{th}$ video.
$\lambda$ is a hyper-parameter balancing the classification loss and the flow loss, where optical flow estimation can be seen as a regularizer for the model to learn motion feature for classification.

Although previous work on multitask learning~\cite{misraSGH16} suggests that sharing parameters of two different tasks may hurt performance, this architecture performs well since optical flow is known empirically to improve video action recognition significantly.
In addition, our architecture contains multiple skip connections from lower convolutional layers to decoder.
This allows higher layers in the encoder to focus on learning more abstract and high level features, without constraining them to remembering all spatial details for predicting optical flow, which is beneficial for action recognition.
This idea is central to Ladder Networks~\cite{rasmusVHBR15} which introduced lateral connections to learn denoising functions and significantly improved classification performance.

It is worth noting that this is a very general architecture and requires minimal architectural engineering.
Thus, it can be trivially extended to learn more tasks jointly to adapt knowledge from different domains.

\noindent\textbf{ActionFlowNet Inference.}
During inference for action classification, optical flow estimation is not required since the motion information is already learned in the encoder. Therefore, the decoder can be removed and only the forward pass of the encoder and the classifier are computed. If the same backbone architecture is used, our model runs at the same speed as a single-stream RGB network without extra computational overhead. Since the optical flow estimation and flow-stream CNN are not needed, it is more efficient than two-stream counterparts.

\subsection{Two-Frame Based Models}
\label{sec:models}
In this section, we propose various models that take two consecutive input frames.
Experimenting with two-frame models has three benefits.
First, when there are multiple frames in the input, it is difficult to determine whether the performance improvement comes from motion modeling or aggregating long term appearance information.
Thus for better analysis, it is desirable to use the two frame input.
Second, training two-frame models is computationally much more efficient than multi-frame models which take $N$ video frames and output $N-1$ optical flow images.
Third, we can measure the effectiveness of external large scale optical flow datasets, such as the FlyingChairs dataset~\cite{FischerDIHHGSCB15}, which provide ground-truth flow on only two consecutive frames, for action recognition.

\noindent\textbf{Learning Optical Flow with ResNet.}
Similarly, we use ResNet-18 as our backbone architecture and learn optical flow.
Like FlowNet-S~\cite{FischerDIHHGSCB15}, we concatenate two consecutive frames to produce a $6(\text{ch})\times224(\text{w})\times224(\text{h})$ input for our two frames model.
At the decoder, there are four outputs with different resolutions.
The total optical flow loss is the weighted sum of end-point error at multiple resolutions per the following equation:
\vspace{-.5em}
\begin{equation}
    \sum_{r=1}^{4} \alpha_r \sum_{p} \|\mathbf{o}_{j,t,p}^{(r)} - \widehat{\mathbf{o}_{j,t,p}^{(r)}} \|_2,
	\label{eq:loss_epe}
\vspace{-.5em}
\end{equation}
%where $\mathbf{o}_{j,t,p}^{(r)}$ is 2-dimensional optical flow vector of the $r^\text{th}$ layer output of the $t^{th}$ and the $(t+1)^{st}$ frame in the $j^{th}$ video at pixel $p$.
where $\mathbf{o}_{j,t,p}^{(r)}$ is the optical flow vector of the $r^\text{th}$ layer output and
$\alpha_r$ is the weighting coefficient of the $r^\text{th}$ optical flow output.
We refer to this pre-trained optical flow estimation network as \emph{FlowNet}.

We first propose an architecture to classify actions on top of the optical flow estimation network, which we call the \emph{Stacked Model}.
Then, we present the two-frame version of ActionFlowNet to classify the actions and estimate the
optical flow, which we call the \emph{ActionFlowNet-2F}.

%With network initialized the flow, the network could exploit the learned motion information to classify actions.
%However, transfering network is known to have the problem of catastrophic forgetting, where the network forgets the previous learned knowledge during finetuning.
%Moreover, the source task optical flow estimation and target task action classification are semantically very different, it is unclear that the knowledge can be effectively transferred by simple finetuning.

%%---------------------------------
\subsubsection{Stacked Model}
\begin{figure}[h!]
\centering
	\includegraphics[width=1.00\linewidth]{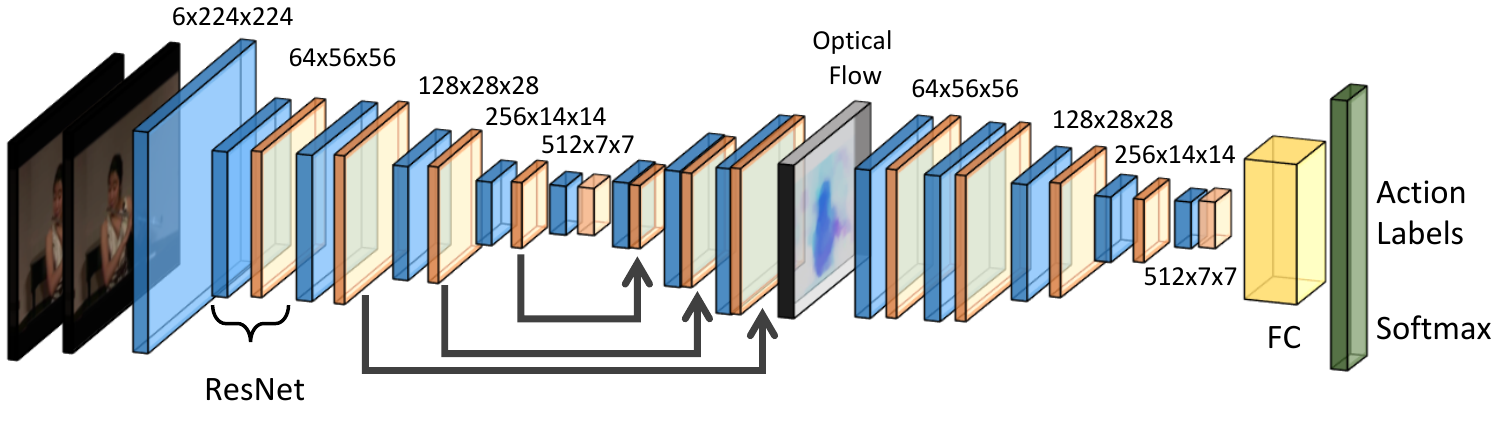}	
  \caption{Network structure of the `Stacked Model'.}
	\label{fig:stacked_archi}
  \vspace{-.5em}
\end{figure}

A straightforward way to use the trained parameters from FlowNet is to take the output of FlowNet and learn a CNN on top of the output, as shown in Figure~\ref{fig:stacked_archi}.
This is reminiscence of the temporal stream in~\cite{simonyanZ14a} which learns a CNN on precomputed optical flow.
If the learned optical flow has high quality, it should give similar performance to learning a network on optical flow.

%Given two frames as input ($t^{th}$ and $(t+1)^{st}$ frame of $j^{th}$ video), the loss function is defined for the first frame of the pair, $t^{th}$ frame, by a standard cross entropy classification loss:
%\begin{equation} 
%	\text{Stacked-Loss}_{j,t} = -\mathbbm{1}(y_j = \widehat{y_j})\log p(\widehat{y_j}),
%\label{eq:stacked_loss}
%\end{equation}
%where $\mathbbm{1}(\cdot)$ is the indicator function. $y_j$ and $p(\widehat{y_j})$ are the action label and predicted probability of class $\widehat{y_j}$ of $j^{th}$ video respectively.

Since the output of FlowNet has 4 times lower resolution than the original image, we remove the first two layers of the CNN (conv1 and pool1) and stack the network on top of it.
We also tried to upsample the flow to the original resolution and use the original architecture including conv1 and pool1, but this produces slightly worse results and is computationally more expensive.

The stacked model introduces about 2x number of parameters compared to the original ResNet, and is also 2x more expensive for inference.
It learns motion features by explicitly including optical flow as an intermediate representation, but cannot model appearance and motion simultaneously, similar to learning a CNN on precomputed optical flow.

\begin{figure}[th!]
\centering
	\includegraphics[width=.80\linewidth]{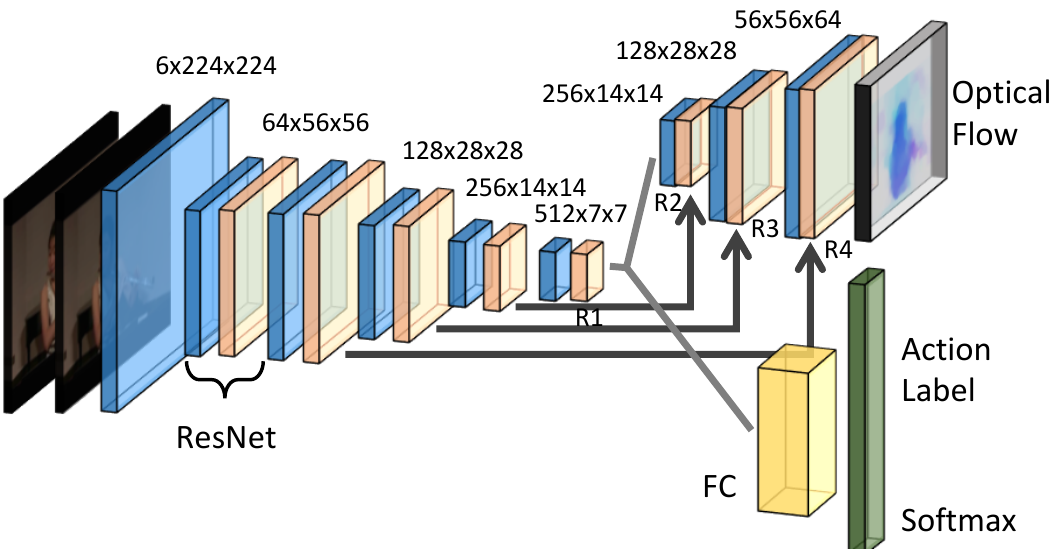}	
	\caption{Network structure of the ActionFlowNet-2F}
	\vspace{-2em}
	\label{fig:mt_archi}
\end{figure}

\subsubsection{ActionFlowNet-2F}
The multitask ActionFlowNet-2F architecture, as illustrated in Figure~\ref{fig:mt_archi}, is based on the two-frame FlowNet with additional classifier. 
Similar to ActionFlowNet, classification is performed by average pooling the last convolutional layer in the encoder followed by a linear classsifier.

Just as with the stacked model, the loss function is defined for each frame.
For the $t^{th}$ frame in the $j^{th}$ video the loss is defined as a weighted sum of classification loss and optical flow loss:
\vspace{-1em}
%\begin{equation} 
\begin{multline} 
%\resizebox{8cm}{!}{$
    \text{MT-Loss}_{j,t} = \underbrace{-\mathbbm{1}(y_j = \widehat{y_j})\log p(\widehat{y_j})}_{\text{Classification Loss}}~~ +\\
    ~~\lambda \underbrace{ \sum_{r=1}^{4} \alpha_r \sum_{p} \|\mathbf{o}_{j,t,p}^{(r)} - \widehat{\mathbf{o}_{j,t,p}^{(r)}} \|_2 }_{\text{Flow Loss}},
%$}
\label{eq:mt_loss}
%\end{equation}
\end{multline}

\vspace{-2.5em}

%!TEX root = ./wacv2018_flowaction.tex

\section{Experiments}
\label{sec:exp}

%%%%%%%%%%%%%%%%%%%%%%%%%%%%%%%%%%%
\subsection{Datasets}
We use two publicly available datasets, UCF101 and HMDB51, to evaluate action classification accuracy.
The UCF101 dataset contains 13,320 videos with 101 action classes~\cite{ucf101}.
The HMDB51 contains 6,766 videos with 51 action categories~\cite{kuehneHETT11}.
As the number of training videos in HMDB51 is small, we initialized our models trained on UCF101 and fine-tuned for HMDB51 similar to~\cite{simonyanZ14a}

The UCF101 and HMDB51 do not have groundtruth optical flow annotation.
Similar to~\cite{tranbftp16}, we use EpicFlow~\cite{ravaudWHS15} as a psuedo-groundtruth optical flow to train the motion part of the network.

To experiment models with better learned the motion signature, we also use FlyingChairs dataset~\cite{FischerDIHHGSCB15} as it has groundtruth optical flow since it is a synthetic dataset.
The FlyingChairs dataset contains 22,872 image pairs and ground truth flow from synthetically generated chairs on real images.
We use the Sintel dataset~\cite{butlerWSB12}, which provides dense groundtruth optical flow, to validate the quality of optical flow models.

%%%%%%%%%%%%%%%%%%%%%%%%%%%%%%%%%%%
\subsection{Experimental Setup}

\noindent \textbf{Overfitting Prevention.} We use different data augmentations on different datasets and tasks.
On the FlyingChairs dataset for optical flow estimation, we augment the data using multi-scale cropping, horizontal flipping, translation and rotation following~\cite{FischerDIHHGSCB15}.
On the UCF101 dataset for optical flow estimation, we use multi-scale cropping and horizontal flipping, but do not use translation and rotation in order to maintain the original optical flow distribution in the data.
On UCF101 dataset for action recognition, we use color jittering~\cite{szegedyLJSRAEVR15}, multi-scale cropping and horizontal flipping.
Dropout is applied to the output of the average pooling layer before the linear classifier with probability 0.5.

\noindent \textbf{Optimization and Evaluation.} The models are trained using Adam~\cite{kingmaB15} for 40,000 iterations with batch size 128 and learning rate $1\times10^{-4}$.
For evaluation, we sample 25 random video segments from a video and run a forward pass to the network on the 10-crops (4 corners + center with their horizontal reflections) and average the prediction scores.% similar to~\cite{simonyanZ14a}.

%%%%%%%%%%%%%%%%%%%%%%%%%%%%%%%%%%%
\subsection{Improving Action Recognition}
We first evaluate the action recognition accuracy by the various proposed two-frame models described in Section~\ref{sec:models}, and then the multi-frame models in Section~\ref{sec:multi}, on both UCF101 and HMDB51 datasets.
All models take RGB inputs only without external optical flow inputs.
%We initialize our finetuning and multitask models with FlowNet trained on FlyingChairs dataset, and FlowNet-Stack model trained on UCF101 dataset.
The recognition accuracies are summarized in Table~\ref{tab:results}.

\begin{table}[h!]
\begin{center}
{\small
    \begin{tabular}{l|c|c}
    \hline
    Method & UCF101 & HMDB51\\
    \hline\hline
      \multicolumn{3}{c}{Two-frame Models} \\
      \hline
    Scratch & 51.3 & 23.9 \\
    FlowNet fine-tune & 66.0 & 29.1 \\
    \hline
    Stacked & 69.6 & 42.4  \\
    ActionFlowNet-2F (UCF101)         & 70.0 & 42.4 \\
    ActionFlowNet-2F (FlCh+UCF101)  & \textbf{71.0} & \textbf{42.6}  \\
    \hline
    ImageNet pretrained ResNet-18 & 80.7 & 47.1\\

    \hline \hline
      \multicolumn{3}{c}{Multi-frame Models} \\
      \hline
    Multi-frame FlowNet fine-tune  & 80.8 & 50.6 \\
    ActionFlowNet (UCF101) & \textbf{83.9} & \textbf{56.4} \\
      \hline
    Sports-1M pretrained C3D~\cite{tranBFTP15} & 82.3 & 53.5\\
    Kinetics pretrained I3D~\cite{carreira2017quo} & 95.6 & 74.8\\
    \hline
    \end{tabular}
}
\end{center}
  \vspace{-1em}
  \caption{Action recognition accuracies of our models on UCF101 and HMDB51 datasets (split 1). FlCh denotes FlyingChairs dataset. ``ActionFlowNet-2F (UCF101)'' denotes its FlowNet part is pretrained on UCF101, and ``ActionFlowNet-2F (FlCh+UCF101)'' denotes its FlowNet part is pretrained on FlyingChairs dataset. All ActionFlowNets are then learned on UCF101 dataset for action and flow. For reference, we additionally show the results trained with large scale datasets~\cite{tranbftp16,carreira2017quo}, but it is not directly comparable since our models are trained with significantly less annotation.}
\label{tab:results}
\end{table}

\textbf{Two-frame Models.}
`Scratch' is a ResNet-18 model that is trained from scratch (random initialization) using UCF101 without any extra supervision, which represents the baseline performance without motion modeling.
`FlowNet fine-tune' is a model that is pretrained from UCF101 for optical flow only, and then fine-tuned with action classification, which captures motion information by initialized FlowNet.
`Stacked' is a stacked classification model on top of optical flow output depicted in Figure~\ref{fig:stacked_archi}. Its underlying FlowNet is trained with UCF101 and is fixed to predict optical flow, so only the CNN classifier on top is learned.
`ActionFlowNet-2F' is the multitask model depicted in Figure~\ref{fig:mt_archi}, which is trained for action recognition and optical flow estimation to learn both motion and appearance.
We trained two versions of ActionFlowNet-2F: one with FlowNet pretrained on UCF101 and one on FlyingChairs dataset.

%All the models compared are well tuned with hyper-parameters. 
As shown in the table, all proposed models - `FlowNet fine-tune', `Stacked' and `ActionFlowNet-2F' significantly outperform `Scratch' .
This implies that our models can take advantage of the learned motion for action recognition, which is difficult to learn implicitly from action labels.

Both the Stacked model and two ActionFlowNet-2Fs outperform the finetuning models by a large margin (up to $5.0\%$ in UCF101 and up to $13.5\%$ in HMDB51).
As all models are pretrained from the high quality optical flow model, the results show that the knowledge learned from previous task is prone to be forgotten when learning new task without multitask learning.
With extra supervision from optical flow estimation, multitask models regularize the action recognition with the effort of learning the motion features.

While the Stacked model performs similarly to ActionFlowNet-2F when trained only on UCF101, ActionFlowNet-2F is much more compact than the Stacked model, containing only approximately half the number of parameters of the Stacked model.
When ActionFlowNet-2F is first pretrained with FlyingChairs, which predicts better quality optical flow in EPE, and finetuned with the UCF101 dataset, it further improves accuracy by $1\%$.
This implies that our multitask model is capable of transferring general motion information from other datasets to improve recognition accuracy further.

Our ActionFlowNet-2F still performs inferior compared to ResNet pretrained on ImageNet, especially in UCF101 (71.0\% vs 80.7\%) because of the rich background context appearance in the dataset. When evaluated on HMDB51, where the backgrounds are less discriminative, our ActionFlowNet-2F is only slightly behind the ImageNet pretrained model (42.6\% vs 47.1\%), indicating that our model learns strong motion features for action recognition.

%Compared to the temporal stream models~\cite{simonyanZ14a}, ActionFlowNet runs significantly faster at test time ($0.005$ second per image in mini-batch) by orders of magnitude in seconds, as optical flow computation (about $0.06$ second per image) is not needed at test time.

\textbf{Multi-frame Models.}
We train 16-frame ActionFlowNet on UCF101. The results are shown in the lower part of Table~\ref{tab:results}. By taking more frames per model, our multi-frame models significantly improve two-frame models (83.9\% vs 70.0\%). This confirms previous work~\cite{karpathyTSLSF14,ngHVVMT15} that taking more input frames in the model is important.

Remarkably, without pretraining on large amounts of labeled data, our ActionFlowNet outperforms the ImageNet pretrained single frame model and Sports-1M pretrained C3D. Our ActionFlowNet gives 1.6\% and 2.9\% improvements over C3D on UCF101 and HMDB51 repsectively. 
The recently published I3D models~\cite{carreira2017quo} achieved strong performance by training on the newly released Kinetics dataset~\cite{kay2017kinetics} with large amount of clean and trimmed labeled video data and performing 3D convolutions on 64 input frames instead of 16 frames.
Although the I3D model achieved better results compared to previous work, their RGB model could still benefit from optical flow inputs, which indicates that even with large amount of labeled data the I3D model does not learn motion features effectively.

It should be noted that there is prior work that gives better results with the use of large scale datasets like ImageNet and Kinetics dataset~\cite{carreira2017quo}, or with the help of external optical flow input~\cite{simonyanZ14a}. Those results are not directly comparable to us because we are using a significantly smaller amount of labeled data - only UCF101 and HMDB51.
Nevertheless, our method shows promising results for learning motion representations from videos. Even with only a small amount of labeled data, our action recognition network outperforms methods trained with a large amount of labeled data with the exception of the recently trained I3D models~\cite{carreira2017quo} which used ImageNet and Kinetics dataset~\cite{kay2017kinetics}.
We envision the performance of ActionFlowNet would further improve when trained on larger datasets like Kinetics and taking more input frames in the model. 
%We take this as future work since training such models on the Kinetics dataset is extremely computationally expensive.

\begin{table}[h!]
  \vspace{-.5em}
\begin{center}
    \begin{tabular}{l|c}
      \hline
      Method & UCF101 Accuracy \\
    \hline\hline
    ResNet-18 Scratch         & 51.3 \\
    VGG-M-2048 Scratch~\cite{simonyanZ14a} & 52.9 \\
%    \hline
    Sequential Verification~\cite{misra2016shuffle} & 50.9 \\
    VGAN~\cite{vondrick2016generating} & 52.1 \\
    O3N~\cite{fernando2016self} & 60.3 \\
    OPN~\cite{lee2017unsupervised} & 59.8 \\
    \hline
    FlowNet fine-tuned (ours) & 66.0 \\
    ActionFlowNet-2F (ours)  & 70.0 \\
    \hline
    ActionFlowNet (ours)  & \textbf{83.9} \\
      \hline
    \end{tabular}
    \vspace{-1em}
\end{center}
  \caption{Results on UCF101 (split 1) from single stream networks with raw pixel input and without pretraining on large labeled dataset.}
  \vspace{-1em}
\label{tab:compare_recent}
\end{table}

\begin{figure*}[h!]
  \vspace{-1em}
\begin{center}
  \begin{subfigure}[ht]{0.18\linewidth}
      \includegraphics[width=0.99\linewidth]{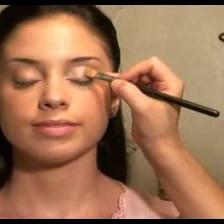}
  \end{subfigure}
  \begin{subfigure}[ht]{0.18\linewidth}
      \includegraphics[width=0.99\linewidth]{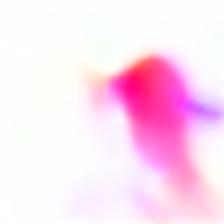}
  \end{subfigure}
  \begin{subfigure}[ht]{0.18\linewidth}
      \includegraphics[width=0.99\linewidth]{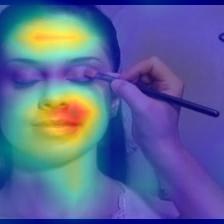}
  \end{subfigure}
  \begin{subfigure}[ht]{0.18\linewidth}
      \includegraphics[width=0.99\linewidth]{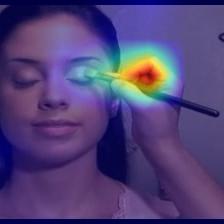}
  \end{subfigure}
  \\
  \begin{subfigure}[ht]{0.18\linewidth}
      \includegraphics[width=0.99\linewidth]{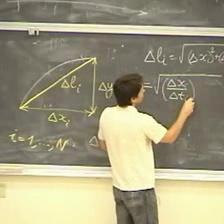}
  \end{subfigure}
  \begin{subfigure}[ht]{0.18\linewidth}
      \includegraphics[width=0.99\linewidth]{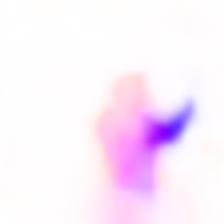}
  \end{subfigure}
  \begin{subfigure}[ht]{0.18\linewidth}
      \includegraphics[width=0.99\linewidth]{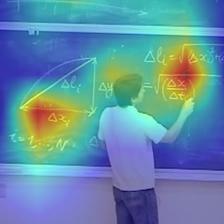}
  \end{subfigure}
  \begin{subfigure}[ht]{0.18\linewidth}
      \includegraphics[width=0.99\linewidth]{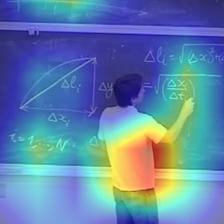}
  \end{subfigure}

  \begin{subfigure}[ht]{0.18\linewidth}
      \caption{Image}
  \end{subfigure}
  \begin{subfigure}[ht]{0.18\linewidth}
      \caption{Output flow from ActionFlowNet-2F}
  \end{subfigure}
  \begin{subfigure}[ht]{0.18\linewidth}
      \caption{ImageNet Model: Appearance Only}
  \end{subfigure}
  \begin{subfigure}[ht]{0.18\linewidth}
      \caption{ActionFlowNet-2F: Motion and Appearance}
  \end{subfigure}
  \vspace{-1em}
  \caption{Visualization of important regions for action recognition. Our ActionFlowNet-2F discovers the regions where the motions are happening to be important while `Appearance Only' captures discriminative regions based on the appearance.}
  \label{fig:qual}
  \vspace{-2em}
\end{center}
\end{figure*}

\textbf{Comparison to state-of-the-arts.}
We compare our approach to previous work that does not perform pretraining with external large labeled datasets in Table~\ref{tab:compare_recent} on UCF101.
All models are trained only with UCF101 labels with different unsupervised learning methods.
Our models significantly outperform previous work that use videos for unsupervised feature learning~\cite{misra2016shuffle, vondrick2016generating, fernando2016self, lee2017unsupervised}.
Specifically, even with only our two-frame fine-tuned model on UCF101, the model obtain more than 5.9\% improvement compared to Sequential Verification, VGAN and O3N, indicating the importance of motion in learning video representations.
When combined with multitask learning, the performance improves to 70.0\%. Finally, when extending our model to 16 frames by 3D convolutions, the performance of ActionFlowNet further boost to 83.9\%, giving a \emph{23.6\% improvement} over the best previous work.
This shows that explicitly learning motion information is important for learning video representations.

%All previous works benchmarked on the UCF101 datasets use external source of data to train their deep model and taking advantage of state-of-the-art optical flow output as input~\cite{tranBFTP15,ngHVVMT15,simonyanZ14a}.
%Thus, it is not fair to compare the performance of our models to them.
%But we compare the performance for the reference purposes in the supplementary material due to page restriction. 

%%=============================
\vspace{-.5em}
\subsubsection{Learning Motions for Discriminative Regions}
\vspace{-.5em}

\begin{figure}
\centering
	\includegraphics[width=.95\linewidth]{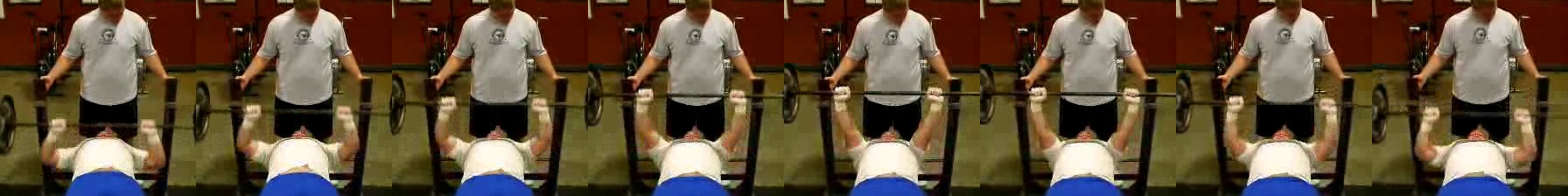}
	\includegraphics[width=.95\linewidth]{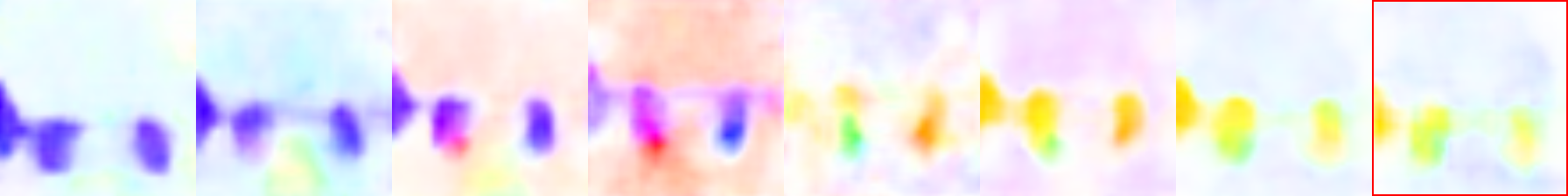}
	\includegraphics[width=.95\linewidth]{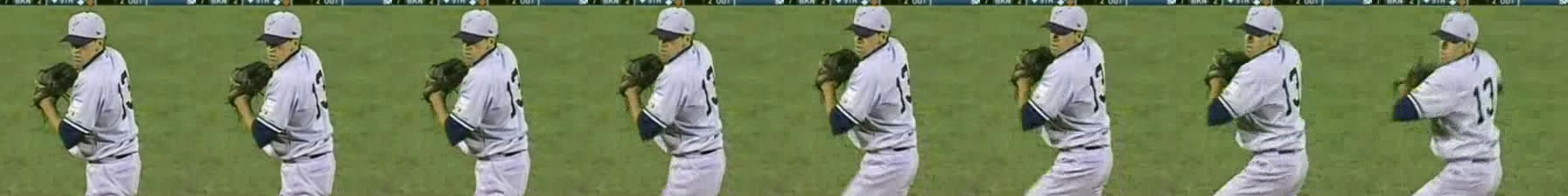}
	\includegraphics[width=.95\linewidth]{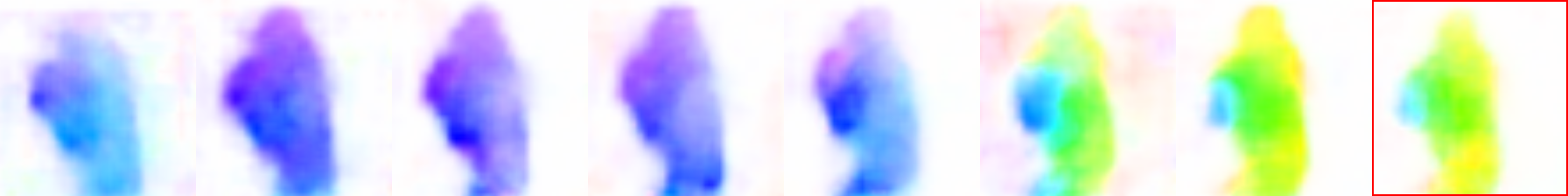}
    \caption{Optical flow and future prediction outputs from our multi-frame model. The 1st and 3rd row shows an example of input videos, and the 2nd and 4th row shows the corresponding optical flow outputs. The last optical flow output frames (in red border) are extrapolated rather than computed within input frames. Only last 8 frames are shown per sample due to space limit.}
	\label{fig:future}
  \vspace{-1em}
\end{figure}

We visualize what is learned from the multitask network by using the method from~\cite{zeilerF14} by using a black square to occlude the frames at different spatial locations and compute the relative difference between classification confidence before and after occlusion. We visualize the two-frame based ActionFlowNet-2F for more straightforward visualization.

We compare the discriminative regions discovered by our multitask network with ones by the ImageNet pretrained ResNet-18, which only models the discriminative appearances without motion.
Figure~\ref{fig:qual} shows example results.
The visualization reveals that our model focuses more on motion, while the ImageNet pretrained network relies more on background appearance, which may not directly relate to the action itself.
However, when appearance is discriminative - for example the writing on the board in the last example - our model can also focus on appearance, which is not possible for models that learn from optical flow only.

\vspace{-.5em}
\subsubsection{Optical Flow and Future Prediction}
\vspace{-.5em}
Figure~\ref{fig:future} shows the optical flow estimation and prediction results from our multi-frame model.
Although the model does not have accurate optical flow groundtruth for training, the optical flow quality is fairly good.
The model predicts reasonable future optical flow, which shows semantic understanding from the model to the frames in addition to simply performing matching between input frames.
\begin{figure}[h!]
\centering
  \begin{subfigure}[ht]{\linewidth}
	\includegraphics[width=\linewidth]{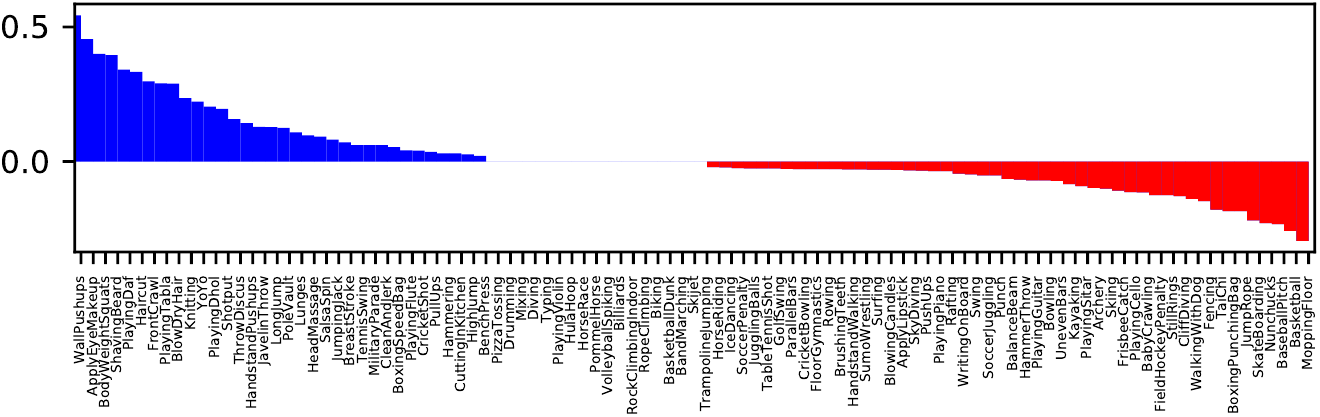}
    \vspace{-1.2em}
    \subcaption{ActionFlowNet vs C3D}
  \end{subfigure}
  \begin{subfigure}[ht]{\linewidth}
	\includegraphics[width=\linewidth]{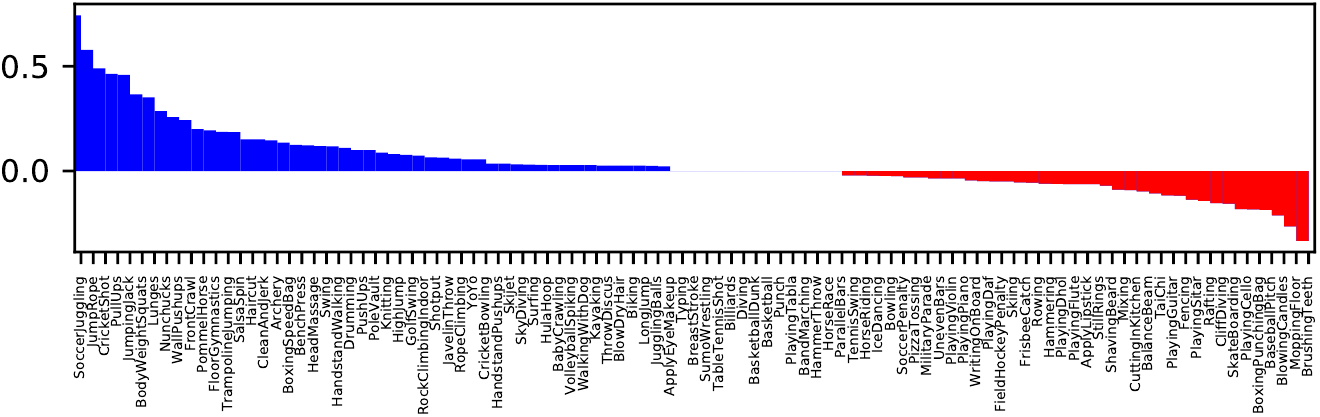}
    \subcaption{ActionFlowNet vs ImageNet pretrained ResNet-18}
  \end{subfigure}
  \vspace{-.5em}
    \caption{Classwise accuracy improvement by ActionFlowNet over pretrained models. The blue bars show positive improvements and the red ones show otherwise.}
	\label{fig:class_wise_improvement}
  \vspace{-1.5em}
\end{figure}

\subsubsection{Classes Improved By Learning Motions}
\vspace{-.5em}
We compare the per class accuracy for ActionFlowNet, ImageNet pretrained model and C3D.
Not all action classes are motion-centric - objects and their contextual (background) appearances provide more discriminative information for some classes~\cite{jain2015what}, which can greatly benefit from large amounts of labeled data.
As shown in Figure~\ref{fig:class_wise_improvement}, our model better recognizes action classes with simple and discriminative motion like WallPushups and ApplyEyeMakeup, while C3D and ImageNet models perform better on classes with complex appearance like MoppingFloor and BaseballPitch.

\subsection{Recognition and Optical Flow Quality}
\label{sec:estimate_of}
\vspace{-.5em}

In this section, we study the effects of different optical flow models for action recognition based on the two-frame models. We train our optical flow models on FlyingChairs or UCF101 and evaluate their accuracies on the Sintel dataset (similar to \cite{FischerDIHHGSCB15} that trains the model on FlyingChairs but tests on other datasets).
%The flow accuracy is evaluated by the end-point-error (EPE) measure, which calculates the averaged $L_2$ distance between the predicted and groundtruth flow vector at each pixel.

We investigate how the quality of the learned optical flow affects action recognition.
Since optical flow in the multitask model is collaboratively learned with the recognition task, the quality of optical flow in the multitask model does not directly affect recognition accuracy.
Thus, we use our Stacked model learned with different datasets, fix the optical flow part and train the classification part in the network shown in Figure~\ref{fig:stacked_archi}.
We compare the end-point-error of different optical flow learners and the corresponding classification accuracy in Table~\ref{tab:epe}.

%With accurate groundtruth labels, the FlowNet trained with FlyingChairs dataset performs quantitatively better than the one trained with UCF101 data: 9.12 vs 11.84 in EPE.
%The flow accuracies from our models are well behind the state of the art optical flow algorithms like EpicFlow, but they are competitive to a widely used optical flow algorithm like TV-L1~\cite{zachPB07}, which gives 10.46 EPE on Sintel dataset.

\begin{table}[h]
  \vspace{-.5em}
\begin{center}
\resizebox{8.0cm}{!}{
  \begin{tabular}{l||c|>{\centering\arraybackslash}m{20mm}}
      \hline
      Method & EPE on Sintel & Classification\newline Accuracy (\%) \\
      \hline\hline
      Stacked on FlyingChairs & \textbf{9.12} & 51.7 \\
      Stacked on UCF101 & 11.84 & \textbf{69.6} \\
      \hline\hline
      ResNet on EpicFlow & 6.29 & 77.7 \\
      \hline
    \end{tabular}
}
\end{center}
  \vspace{-1em}
  \caption{Comparison between End-Point-Error (EPE, lower is better) and the classification accuracy. Interestingly, better optical flow does not always result in better action recognition accuracy. Refer to the text for discussion.}
\label{tab:epe}
  \vspace{-.5em}
\end{table}

\begin{figure}[h!]
\begin{center}
  \begin{subfigure}[ht]{0.45\linewidth}
      \includegraphics[width=0.99\linewidth]{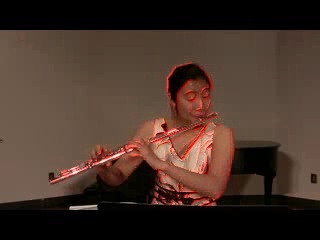}
  \end{subfigure} 
  \begin{subfigure}[ht]{0.45\linewidth}
      \includegraphics[width=0.99\linewidth]{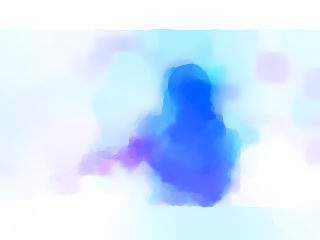}
  \end{subfigure}
  \\
  \begin{subfigure}[ht]{0.45\linewidth}
      \caption{Frame with movements highlighted with red}
  \end{subfigure} 
  \begin{subfigure}[ht]{0.45\linewidth}
      \caption{EpicFlow}
  \end{subfigure}
  \vspace{-.5em}
  \\
  \begin{subfigure}[ht]{0.45\linewidth}
      \includegraphics[width=0.99\linewidth]{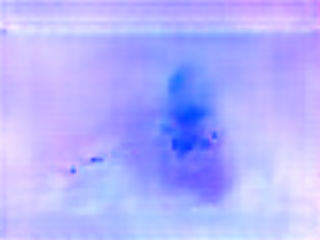}
  \end{subfigure}
  \begin{subfigure}[ht]{0.45\linewidth}
      \includegraphics[width=0.99\linewidth]{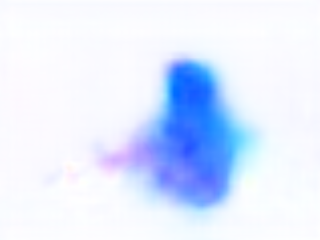}
  \end{subfigure}
  \\
  \begin{subfigure}[ht]{0.45\linewidth}
      \caption{FlowNet on FlyingChairs}
  \end{subfigure}
  \begin{subfigure}[ht]{0.45\linewidth}
      \caption{FlowNet on UCF101}
  \end{subfigure}

  \vspace{-1em}
  \caption{Qualitative comparison of flow outputs. It shows an example of small motion, where the maximum magnitude of displacement estimated from EpicFlow is only about 1.6px. FlowNet trained on FlyingChairs dataset fails to estimate small motion, since the FlyingChairs dataset consists of large displacement flow.}
  \label{fig:flow-example}
  \vspace{-1.5em}
\end{center}
\end{figure}

%-- action recognition
\noindent\textbf{Action Recognition with Learned Flow.}
Surprisingly, even with lower end-point-error the Stacked model pretrained on FlyingChairs performs significantly worse than the one pretrained on UCF101 dataset (51.7\% vs 69.6\%), as shown in Table~\ref{tab:epe}.
Compared to the model directly taking high quality optical flow as input ($77.7\%$), our models are still not as good as training directly on optical flow.
We believe this is because the quality of learned optical flow is not high enough.

To understand how the learned optical flow affects action recognition, we qualitatively observe the optical flow performance in Figure~\ref{fig:flow-example}.
Even though the end-point error on Sintel of the FlowNet pretrained on FlyingChairs is low, the estimated optical flow has lots of artifacts in the background and the recognition accuracy on top of that is correspondingly low.
We believe the reason is that the FlyingChairs dataset mostly consists of large displacement flow, and therefore the model performs badly on estimating small optical flow, which contributes less in the EPE metric when averaged over the whole dataset.
This is in contrast to traditional optimization based optical flow algorithms that can predict small displacements well but have difficulties for large displacements.

In addition, traditional optical flow algorithms such as TV-L1 and EpicFlow explicitly enforce smoothness and constancy. They are able to preserve object shape information when the flow displacements are small, which is important for action recognition.
While our models perform comparably to traditional optical flow algorithms in terms of endpoint error, our model is not optimized for preserving flow smoothness. 
This shows that end-point-error of optical flow in public dataset may not be a good indicator of action classification performance, since shape preservation is not accounted for in the metric.

%We additionally trained the UCF101 FlowNet-Stack model in an end-to-end manner, where the pretrained FlowNet at the bottom is fine-tuned while training the CNN on top.
%We find that the model performs significantly worse, with 56.2\% accuracy because of overfitting and also forgetting the previously learned knowledge - this reflects the difficulty of learning motion features directly from raw frames.

%\begin{table}[h!]
%\begin{center}
%    \begin{tabular}{l|c}
%    \hline
%    Method & Accuracy \\
%    \hline\hline
%    FlowNet-Stack trained on FlyingChairs & 51.7  \\
%    %FlowNet-Stack trained on FlyingChairs (ft) & 52.0  \\
%    FlowNet-Stack trained on UCF101 & 69.6 \\
%    %FlowNet-Stack trained on UCF101 (ft) & 56.2 \\
%    \hline
%    EpicFlow & 77.7 \\
%    \hline
%    \end{tabular}
%    \end{center}
%    \caption{Comparison of flow method of UCF101 split 1}
%    \label{tab:flownet-stack-compare}
%\end{table}

%Our stack model performs inferior to the model directly trained on optical flow.

%!TEX root = ./nips_2016_flowaction.tex

\vspace{-.5em}
\section{Conclusion}
\vspace{-.5em}
\label{sec:conclusion}

We presented a multitask framework for learning action with motion flow, named \emph{ActionFlowNet}. 
By using optical flow as supervision for classification, our model captures motion information while not requiring explicit optical flow computation as input.
Our model significantly outperforms previous feature learning methods trained without external large scale data and additional optical flow input.
%, and achieves competitive recognition accuracy to the models trained with large labeled datasets.
%Our work provide important insights on using optical flow to learn motion representation.

\subsubsection*{Acknowledgments}
This research was supported in part by funds provided from
the Office of Naval Research under grant N000141612713
entitled ``Visual Common Sense Reasoning for Multi-agent
Activity Prediction and Recognition''.

{\small
\bibliographystyle{ieee}
\bibliography{manual}
}

\end{document}